# Reflection-Aware Reasoning for Non-Line-of-Sight Pedestrian Localization

Byeonggyu Park[1], Mingu Jeon[2,*], and Seong-Woo Kim[1,*]

[1] Seoul National University, Seoul, Republic of Korea
{bg.park, snwoo}@snu.ac.kr
[2] Pusan National University, Busan, Republic of Korea
mingujeon@pusan.ac.kr
[*]Corresponding authors.

**Abstract.** Reliable localization of non-line-of-sight (NLOS) pedestrians is critical for safe urban autonomous driving, yet it remains highly challenging in ego-dynamic outdoor environments, where ego-vehicle motion makes radar multipath propagation complex and noisy. In this paper, we present a reflection-aware framework for NLOS pedestrian localization with a moving ego-vehicle in outdoor testbed scenarios. Our framework fuses front-view camera images and 2D radar point clouds to infer reflection orders and reflective surface distributions in bird's-eye-view space. It then uses physics-guided ray tracing to reconstruct distorted reflection paths and localize the hidden pedestrian. We validate the framework in outdoor testbed scenarios under ego-dynamic conditions. The results demonstrate the effectiveness of the proposed framework for NLOS pedestrian localization with a moving ego-vehicle.

## 1 Introduction

Urban driving environments contain dense structural occlusions caused by buildings, walls, and noise barriers. Such structures create non-line-of-sight (NLOS) regions where pedestrians are fully hidden from direct observation. In urban environments, sudden pedestrian emergence from occluded areas poses a persistent safety challenge. In 2023, approximately 84% of pedestrian fatalities in the United States occurred in urban areas, and pedestrian deaths accounted for about 18% of all traffic fatalities [22, 26]. The prevalence of fatalities in structurally complex urban settings underscores the need for perception beyond direct line-of-sight (LOS), as illustrated in Fig. 1.

To prevent accidents caused by pedestrians emerging from occluded regions, recent studies have explored methods for detecting objects in NLOS areas. Representative approaches include vehicle-to-everything (V2X) systems [12, 13, 18], reflective wave-based sensing modalities such as acoustic sensing [9, 28] and mmWave radar [5,10,11,24]. V2X-based methods can provide information about hidden objects through communication, but require prior infrastructure deployment [8]. Sound-based NLOS perception leverages diffraction and reflection properties of acoustic waves [1, 21, 28], yet often assumes predefined spatial layouts and remains sensitive to environmental noise, making robust localization in urban driving environments challenging. mmWave radar-based methods exploit multipath reflections to estimate occluded objects [29, 37]. These approaches typically rely on predefined geometric assumptions or rule-based modeling to interpret reflection paths [10, 37] and are commonly validated in controlled indoor environments [17,36]. Consequently, most existing NLOS perception methods [1, 5, 9–11, 24, 29, 37] assume static sensing setups and stable propagation patterns, which limits their applicability in outdoor ego-dynamic scenarios.

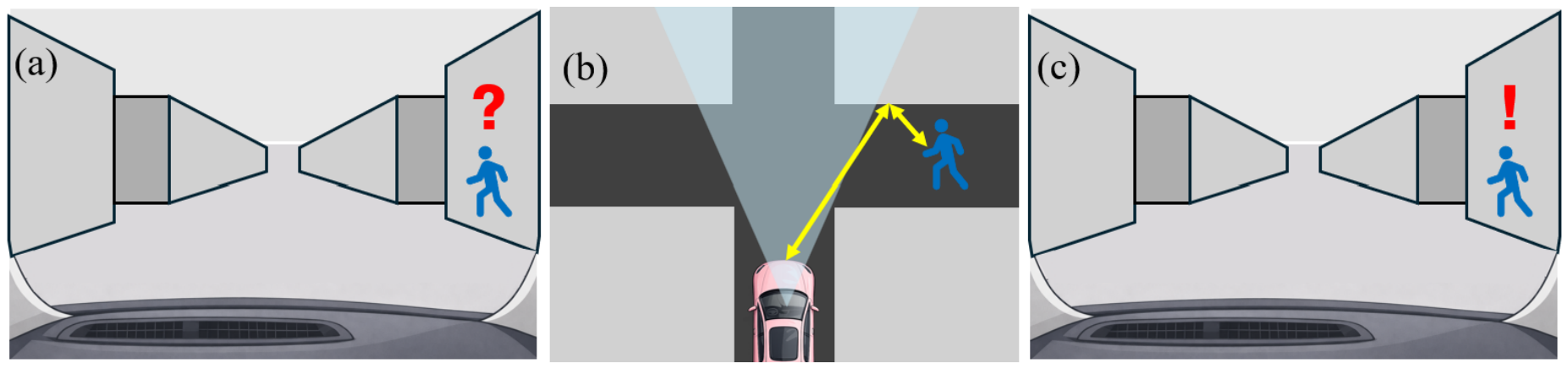

**Fig. 1: Illustration of an NLOS pedestrian scenario from the ego-vehicle perspective.** (a) While driving through a narrow alley, regions between buildings fall outside the LOS, making direct detection of pedestrians difficult. (b) By leveraging reflective wave propagation, indirect spatial evidence can be obtained even when the target is occluded. (c) Based on reflection-aware reasoning, the position of the NLOS object can be inferred despite the absence of direct visibility.

Achieving reliable NLOS perception with radar point cloud data (PCD) in driving environments presents several challenges. Under ego-dynamic conditions, radar observations vary across frames due to vehicle motion and are often corrupted by measurement noise and clutter. Therefore, identifying radar points that originate from NLOS pedestrians becomes challenging. In addition, observations of NLOS pedestrians appear spatially distorted because radar signals propagate through multi-bounce reflection paths. As a result, accurate estimation of reflective surfaces is required to reconstruct reflection paths and localize occluded objects.

To address these challenges, we propose a practical reflection-aware learning framework. The proposed framework consists of two stages: Reflection-aware representation learning and Physics-guided ray tracing. In the reflection-aware representation learning stage, feature-level fusion is performed between front-view camera images providing visual cues and semantic information in the LOS region and 2D radar PCD containing quantitative range measurements and reflection characteristics. Based on the fused representations, the model performs reflection-type point segmentation to classify radar points according to their reflection mechanisms. In addition, reflective surface distributions are estimated as heatmaps in bird's-eye-view (BEV) space. In the physics-guided ray tracing stage, the estimated reflective structures and the identified radar points are used to interpret reflection paths and reconstruct physically propagation trajectories.

**Table 1: Comparison of representative NLOS perception and localization studies**. Existing approaches often rely on predefined geometry or rule-based reasoning and are mostly evaluated in static or controlled settings. In contrast, the proposed method performs learning-based NLOS point extraction and reflective surface estimation, and is validated under ego-dynamic conditions.

| Method | Sensors | Methodology | | Validation Setting | | | |
|---|---|---|---|---|---|---|---|
| | | Learned Points | Learned Surface | Outdoor | Vehicle Val. | Multi-Target | Ego Motion |
| [9] | Mic Array | | | ✓ | ✓ | | |
| [37] | Radar | | | ✓ | | ✓ | |
| [10] | Radar | | | ✓ | ✓ | ✓ | |
| [16] | Radar | ✓ | ✓ | | | | ✓ |
| [31, 32] | Radar | | | | ✓ | | |
| [27] | Radar, LiDAR | ✓ | | ✓ | ✓ | | |
| [11, 24] | Radar, Camera | | ✓ | ✓ | ✓ | ✓ | |
| **Proposed** | Radar, Camera | ✓ | ✓ | ✓ | ✓ | ✓ | ✓ |

To evaluate the effectiveness of the proposed method, experiments are conducted in a constructed outdoor T-junction intersection testbed. The evaluation is performed under both ego-static and ego-dynamic conditions using the average Euclidean localization error. The proposed method achieves localization errors of 0.44 m and 1.01 m for LOS and NLOS pedestrians under ego-static conditions, and 0.54 m and 1.23 m under ego-dynamic conditions, outperforming existing approaches [24, 27]. The results demonstrate reliable localization of NLOS pedestrians even under ego-dynamic driving conditions.

The main contributions of this paper are as follows:

- We propose a practical reflection-aware framework for NLOS pedestrian localization with a moving ego-vehicle in outdoor testbed scenarios.
- We develop a reflection-aware method that combines multimodal representation learning and physics-guided ray tracing to recover distorted radar propagation and localize hidden pedestrians.
- We present experimental validation of the proposed method with a moving ego-vehicle in a outdoor testbed intersection.

Code, data, and models are available on the project page.

## 2 Related Work

*Acoustic-based NLOS perception.* Several approaches have leveraged acoustic waves for non-line-of-sight perception by exploiting their reflection and diffraction properties [1, 2, 21, 28]. Prior work has estimated object directions using microphone arrays [2] or inferred positions by analyzing acoustic reflection paths [1–3]. However, these approaches are primarily evaluated in indoor or static environments with limited noise, restricting applicability to urban driving scenarios.

In outdoor settings, sound-based methods have been explored for vehicle state classification [9] and NLOS vehicle localization using particle filters [7]. Nevertheless, these approaches either focus on coarse state inference or rely on predefined geometric maps, limiting their ability to infer spatial structure directly from acoustic observations in dynamic environments.

*Radar-based NLOS perception.* Unlike acoustic waves, mmWave radar provides accurate range measurements, enabling indirect observation of occluded objects through multipath reflections [29, 32, 37]. Early studies explored around-the-corner radar sensing using geometric reflection modeling and multipath exploitation [31,32]. Later works further investigated localization methods based on multipath radar signals and joint estimation of propagation parameters [37]. While effective in controlled settings, such approaches often rely on predefined wall or corner geometry and prior scene knowledge, limiting their applicability to unknown or dynamic urban environments. Learning-based methods have also been introduced to analyze radar point clouds for NLOS perception. These approaches classify radar points into reflection categories or detect NLOS-associated points using data-driven models [15,27]. More recent studies have explored high-resolution mmWave imaging and RF sensing techniques to reconstruct occluded objects behind reflectors [5, 17]. However, such systems are typically limited to indoor or short-range scenarios with static targets. Recent radar localization studies continue to exploit multipath propagation for NLOS target positioning [4, 33]. Nevertheless, many of these methods rely on auxiliary structural information or controlled environments, limiting their applicability to outdoor driving scenarios.

*Camera–radar fusion for NLOS perception.* To compensate for sparse and noisy radar measurements, several studies have explored camera–radar fusion for object detection and scene understanding, where visual semantic cues complement radar observations [6, 20, 35]. However, these approaches are primarily designed for LOS perception and do not explicitly address multipath effects in NLOS scenarios. For NLOS settings, prior work has applied particle filtering to detect occluded pedestrians [23] or leveraged camera-derived structural cues, such as road layout and reflective surfaces, to guide radar interpretation [11, 24]. Nevertheless, radar point interpretation remains largely based on predefined heuristics rather than data-driven modeling of multipath propagation.

Table 1 summarizes representative NLOS perception and localization studies with respect to their methodological components and validation settings. Prior studies have made substantial progress in around-corner sensing, target localization, and radar-based reconstruction; however, they are predominantly evaluated in static, controlled, or ego-static scenarios. In contrast, our paper addresses outdoor ego-dynamic NLOS pedestrian localization using a moving vehicle-mounted radar-camera platform, including multi-target localization scenarios.

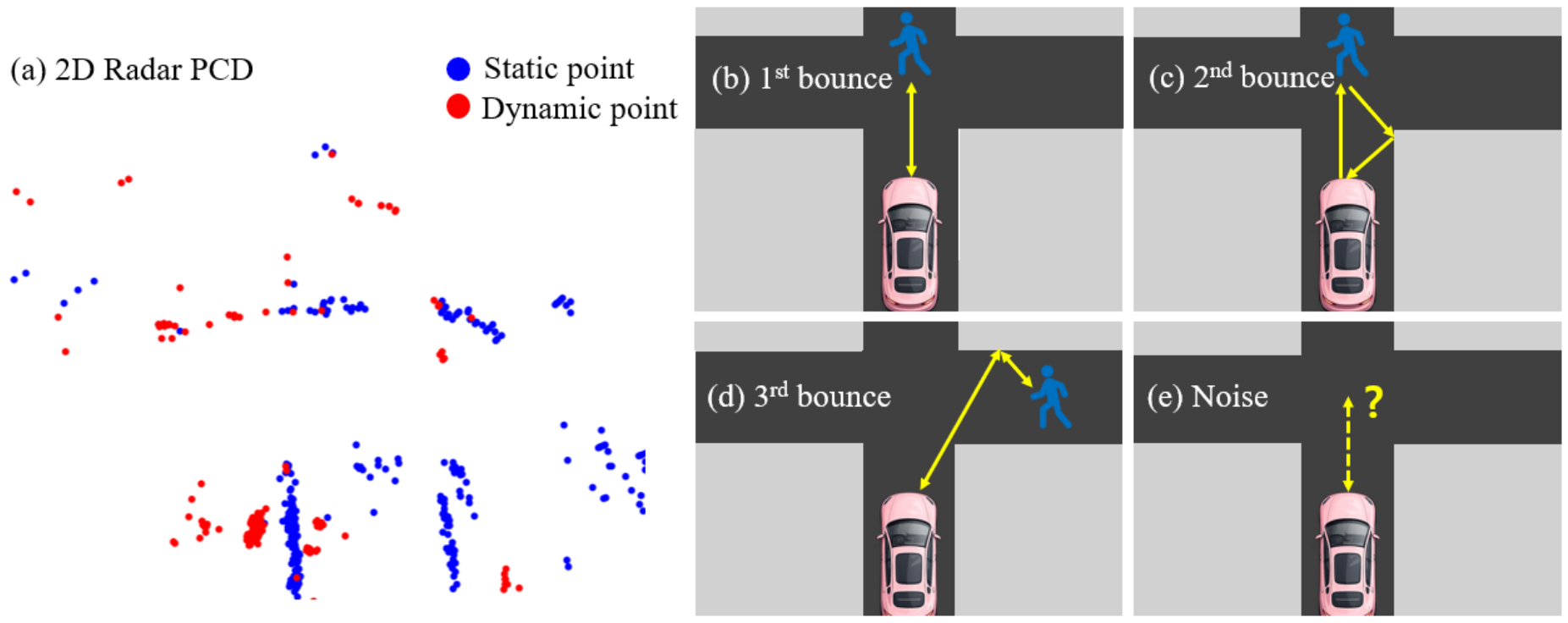

**Fig. 2: Reflection path diversity in mmWave radar observations.** (a) mmWave 2D Radar PCD is observed in BEV and appears sparse and noisy, making informative observations difficult to identify. (b)–(d) illustrate representative multipath propagation cases (1st, 2nd, and 3rd bounce), where the geometric strategy for object localization varies depending on the reflection order. Although higher-order reflections may exist in practice, we model up to third-order reflections considering mmWave attenuation characteristics. (e) shows structurally inconsistent noise that further complicates reflection path interpretation.

## 3 Problem Definition

To formally describe the reflection reasoning problem addressed in this work, we first define the radar reflection path representation used for NLOS interpretation.

*Reflection Path Definition.* Each radar point corresponds to a propagation path characterized by the number of reflections the signal undergoes before returning to the sensor. In this work, we consider up to third-order reflections, taking into account the attenuation characteristics of mmWave signals. Accordingly, each radar point is categorized into first, second, or third-order reflection paths. Furthermore, depending on whether a dynamic object is involved along the propagation path, each point is additionally classified as either object-originated or surface-originated reflection Fig. 2. This labeling scheme jointly encodes the reflection order and the physical origin of the signal, resulting in a structured reflection-path representation.

*Input Representation.* We consider a time-discrete sequence of synchronized sensor measurements indexed by $t \in \{1, \dots, T\}$ where $T$ denotes the total number of time steps in the sequence. At each time step $t$, the ego-vehicle observes a front-view RGB image $\mathbf{I}_t$ and a 2D Radar PCD $\mathbf{R}_t = \{\mathbf{r}_{t,i}\}_{i=1}^{N_t}$.

Each radar point is represented as

$$\mathbf{r}_{t,i} = (x_{t,i}, y_{t,i}), v_{t,i}, \rho_{t,i},$$

where $(x_{t,i}, y_{t,i})$ denotes planar coordinates in the ego-centric BEV frame, $v_{t,i}$ is the Doppler velocity, and $\rho_{t,i}$ represents the radar cross section (RCS). The index $i$ identifies an individual radar point at time $t$, and $N_t$ denotes the number of radar points in $\mathbf{R}_t$, which varies across frames.

*Problem Formulation.* Given synchronized inputs $(\mathbf{I}_t, \mathbf{R}_t)$ at time $t$, the proposed model predicts two complementary intermediate representations: (i) point-wise reflection-path labels $\hat{\mathbf{Y}}_t = \{\hat{y}_{t,i}\}_{i=1}^{N_t}$ and (ii) a reflective surface probability map in BEV space $\hat{\mathbf{S}}_t \in [0,1]^{H_{\text{bev}} \times W_{\text{bev}}}$.

Let $f_\theta$ denote the proposed reflection-aware model parameterized by $\theta$. The model is formally defined as

$$f_\theta : (\mathbf{I}_t, \mathbf{R}_t) \rightarrow (\hat{\mathbf{Y}}_t, \hat{\mathbf{S}}_t). \tag{1}$$

$\hat{\mathbf{Y}}_t$ and $\hat{\mathbf{S}}_t$ are combined through a physics-constrained NLOS inference operator $\mathcal{T}(\cdot)$:

$$\hat{\mathbf{X}}_t = \mathcal{T}(\hat{\mathbf{Y}}_t, \hat{\mathbf{S}}_t), \tag{2}$$

The final objective is to minimize the localization error $e_t$ between the predicted object position $\hat{\mathbf{X}}_t$ and the ground-truth position $\mathbf{X}_{\text{gt},t}$:

$$e_t = \left\| \hat{\mathbf{X}}_t - \mathbf{X}_{\text{gt},t} \right\|_2. \tag{3}$$

# 4 Method

## 4.1 Overview

To achieve this, we develop a learning-based framework that performs feature-level fusion between front-view camera images, which provide structural and semantic cues in LOS regions, and radar point clouds containing quantitative range measurements and reflection characteristics Fig. 3. The model predicts reflection-type labels for radar points and estimates reflective surface distributions in BEV space, which are then integrated into a physics-guided ray tracing module to interpret reflection paths and infer the final NLOS pedestrian location.

## 4.2 Reflection-Aware Representation Learning

While cameras provide geometric and visual layout of the scene, radar offers physical observations of NLOS objects through multipath propagation. To preserve the complementary properties of both modalities, we perform feature-level fusion to learn reflection-aware representations for NLOS pedestrian localization.

The input front-view image $\mathbf{I}_t$ is transformed into BEV space using an Lift-Splat-Shoot (LSS)-based image encoder $f_{\text{cam}}$ [25]. The LSS encoder predicts per-pixel depth distributions and projects image features onto the BEV plane, generating a spatially structured visual representation:

$$\mathbf{F}_{\text{cam}} = f_{\text{cam}}(\mathbf{I}_t), \qquad \mathbf{F}_{\text{cam}} \in \mathbb{R}^{C \times H_{\text{bev}} \times W_{\text{bev}}}. \tag{4}$$

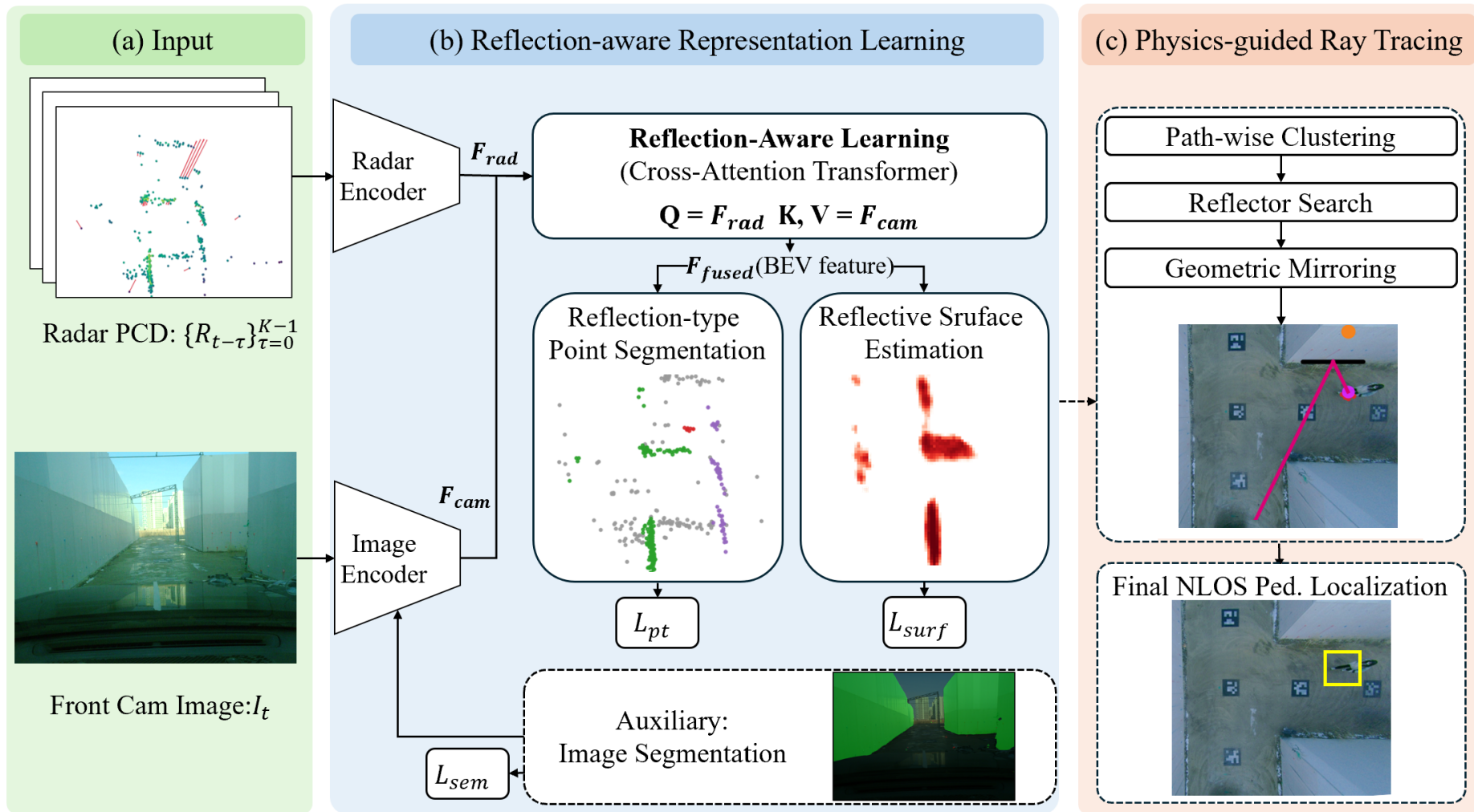

**Fig. 3: Overall framework of the proposed reflection-aware NLOS localization method.** (a) **Input.** A temporal stack of radar point clouds $\{R_{t-\tau}\}_{\tau=0}^{K-1}$ and a synchronized front-view image $I$. (b) **Reflection-aware representation learning.** Radar point features $F_{\text{rad}}$ and BEV-aligned visual features $F_{\text{cam}}$ are extracted and fused via cross-attention to obtain $F_{\text{fused}}$, from which reflection-type point labels and a reflective surface heatmap are predicted. (c) **Physics-guided ray tracing.** Reflection-consistent clusters are formed, candidate reflectors are identified, and the NLOS pedestrian position is recovered via geometric mirroring.

Here, $C$ denotes the feature channel dimension, and $H_{\text{bev}}$, $W_{\text{bev}}$ denote the spatial resolution of the BEV feature map.

To stabilize semantic feature learning, an auxiliary semantic segmentation head is attached to the image encoder during training. The auxiliary loss is defined as a cross-entropy term $\mathcal{L}_{\text{sem}}$. This head is used only during training and discarded at inference time.

For radar, the PCD observed at a single time step $t$ is extremely sparse and noisy. To mitigate this limitation, we aggregate radar observations from the past $K$ frames. Specifically, $\{\mathbf{R}_{t-\tau}\}_{\tau=0}^{K-1}$ are ego-motion compensated and aligned to the current coordinate frame before accumulation. The aggregated radar points are encoded in BEV space using a radar encoder $f_{\text{rad}}$:

$$\mathbf{F}_{\text{rad}} = f_{\text{rad}}\left(\{\mathbf{R}_{t-\tau}\}_{\tau=0}^{K-1}\right), \qquad \mathbf{F}_{\text{rad}} \in \mathbb{R}^{C\times H_{\text{bev}}\times W_{\text{bev}}}. \tag{5}$$

The radar BEV feature $\mathbf{F}_{\text{rad}}$ and camera BEV feature $\mathbf{F}_{\text{cam}}$ are fused via a transformer-based cross-attention module. To allow sparse radar features to reference dense spatial structures, we set radar features as queries and camera features as keys and values:

$$\mathbf{Q} = \mathbf{W}_Q\mathbf{F}_{\text{rad}}, \quad \mathbf{K} = \mathbf{W}_K\mathbf{F}_{\text{cam}}, \quad \mathbf{V} = \mathbf{W}_V\mathbf{F}_{\text{cam}}. \tag{6}$$

The fused representation is then obtained as

$$\mathbf{F}_{\text{fused}} = \text{Softmax}\left(\frac{\mathbf{Q}\mathbf{K}^{\top}}{\sqrt{d}}\right)\mathbf{V}. \tag{7}$$

The resulting fused representation $\mathbf{F}_{\text{fused}}$ jointly encodes the physical reflection characteristics captured by radar and the structured spatial cues provided by the camera. This reflection-aware feature representation serves as the shared input for both the reflection-type point segmentation and reflective surface estimation modules.

### 4.3 Reflection-Type Point Segmentation

To interpret multipath propagation, we classify the reflection order of each radar point in $\mathbf{R}_t$ at the point level. Specifically, the BEV feature vector corresponding to the spatial location of $\mathbf{r}_{t,i}$ is sampled from $\mathbf{F}_{\text{fused}}$ and passed through a point segmentation head $g_\theta$ to predict the reflection-type logits $\mathbf{p}_{t,i}$:

$$\mathbf{p}_{t,i} = g_\theta\left(\mathbf{F}_{\text{fused}}(\mathbf{r}_{t,i})\right), \qquad \mathbf{p}_{t,i} \in \mathbb{R}^{|\mathcal{C}|}, \tag{8}$$

where $\mathcal{C}$ denotes the set of reflection-type classes corresponding to different reflection orders. The predicted reflection label is obtained as

$$\hat{y}_{t,i} = \arg\max_{c\in\mathcal{C}} \mathbf{p}_{t,i}^{(c)}. \tag{9}$$

The point segmentation head is applied independently to each radar point and trained as a multi-class classification problem using a point-wise cross-entropy loss $\ell_{\text{CE}}$:

$$\mathcal{L}_{\text{pt}} = \frac{1}{N_t}\sum_{i=1}^{N_t} \ell_{\text{CE}}\left(\mathbf{p}_{t,i}, y_{t,i}\right), \tag{10}$$

where $y_{t,i}$ is the ground-truth reflection-type label for point $\mathbf{r}_{t,i}$, and $N_t$ denotes the number of radar points at time $t$.

Through this process, informative observations associated with NLOS pedestrians are identified from sparse and noisy $\mathbf{R}_t$ at the reflection-order level. This decomposition separates mixed multipath signals into physically interpretable reflection paths, which subsequently serve as inputs to reflective surface estimation and physics-based NLOS reconstruction.

### 4.4 Reflective Surface Estimation

Although point segmentation identifies higher-order reflection points, recovering the true location of an NLOS pedestrian requires additional geometric constraints that describe the structures responsible for the observed reflections. Since NLOS observations arise from reflections on surrounding environmental structures rather than direct object returns, the location and shape of reflective surfaces must be estimated to interpret the underlying propagation paths.

To estimate reflective structures responsible for multipath propagation, the fused BEV representation $\mathbf{F}_{\text{fused},t}$ is used to predict a reflective surface probability map at time step $t$:

$$\hat{\mathbf{S}}_t = h_\phi(\mathbf{F}_{\text{fused},t}), \qquad \hat{\mathbf{S}}_t \in [0,1]^{H_{\text{bev}} \times W_{\text{bev}}}, \tag{11}$$

where $h_\phi$ denotes the reflective surface prediction head parameterized by $\phi$. Let $(u, v)$ denote a grid coordinate in BEV space. Each element $\hat{\mathbf{S}}_t(u, v)$ represents the probability that a reflective surface exists at location $(u, v)$.

To supervise reflective surface estimation, we construct a pseudo ground-truth surface map $\mathbf{S}_t$. Because radar PCD from a single frame are highly sparse, radar observations from temporally adjacent frames are aggregated after ego-motion compensation. Specifically, radar PCD from $K_h$ past frames and $K_f$ future frames relative to time step $t$ are aligned to the coordinate frame at time $t$. The aggregated radar point set $\tilde{\mathbf{R}}_t$ is defined as

$$\tilde{\mathbf{R}}_t = \bigcup_{\tau=-K_h}^{K_f} \mathbf{R}_{t+\tau}. \tag{12}$$

Among the aggregated radar points, only those classified as first-order surface reflections are retained to form a candidate reflective surface set $\mathbf{R}_t^{\text{surf}}$. Each selected point is then projected onto the BEV grid using anisotropic Gaussian splatting to generate the pseudo surface map:

$$\mathbf{S}_t(u, v) = \max_j \exp\left(-\frac{1}{2}\left[\frac{d_{\parallel,j}^2}{\sigma_\parallel^2} + \frac{d_{\perp,j}^2}{\sigma_\perp^2}\right]\right), \tag{13}$$

where $(u, v)$ denotes a BEV grid location. The terms $d_{\parallel,j}$ and $d_{\perp,j}$ represent the distances from grid cell $(u, v)$ to point $\mathbf{r}_j$ along the estimated tangent and normal directions of the reflective surface, respectively. The parameters $\sigma_\parallel$ and $\sigma_\perp$ control anisotropic smoothing, with $\sigma_\parallel > \sigma_\perp$ to preserve elongated reflective structures while suppressing diffusion across the normal direction.

Reflective surface estimation is formulated as a binary classification problem over BEV grid cells and optimized using the binary cross-entropy loss $\ell_{\text{BCE}}$:

$$\mathcal{L}_{\text{surf}} = \ell_{\text{BCE}}(\hat{\mathbf{S}}_t, \mathbf{S}_t). \tag{14}$$

### 4.5 Physics-Guided Ray Tracing

In this stage, $\hat{y}_{t,i}$ and $\hat{\mathbf{S}}_t$ are used to recover the location of NLOS pedestrians through physics-based ray tracing. The module operates as a deterministic inference procedure grounded in geometric optics and reflection physics. It consists of three steps: (i) path-wise clustering, (ii) ray-guided reflector search, and (iii) geometric mirroring.

*Path-wise Clustering.* Given the $\mathbf{R}_t$, radar points are first separated according to $\hat{y}_{t,i}$. The first-order reflection point set $\mathcal{P}_t^{(1)}$ and third-order reflection point set $\mathcal{P}_t^{(3)}$ are defined as

$$\mathcal{P}_t^{(1)} = \{\mathbf{r}_{t,i} \mid \hat{y}_{t,i} = 1\}, \quad \mathcal{P}_t^{(3)} = \{\mathbf{r}_{t,i} \mid \hat{y}_{t,i} = 3\}. \tag{15}$$

Second-order reflections are excluded from geometric inference because they typically occur under LOS conditions and are accompanied by more reliable first-order reflections.

DBSCAN clustering is then applied to $\mathcal{P}_t^{(1)}$ and $\mathcal{P}_t^{(3)}$ to obtain representative cluster centers, denoted as $\{\bar{\mathbf{r}}_k^{(1)}\}$ and $\{\bar{\mathbf{r}}_m^{(3)}\}$, respectively. The centers $\bar{\mathbf{r}}_k^{(1)}$ approximate the true object location, while $\bar{\mathbf{r}}_m^{(3)}$ correspond to higher-order reflections that do not coincide with the actual object position.

*Ray-Guided Reflector Search.* Each third-order cluster center $\bar{\mathbf{r}}_m^{(3)}$ is assumed to originate from reflections on surrounding structures. Therefore, the corresponding reflective surface must lie along the ray connecting the radar origin $\mathbf{o}$ and $\bar{\mathbf{r}}_m^{(3)}$. The unit ego-ray direction is defined as

$$\mathbf{d}_m = \frac{\bar{\mathbf{r}}_m^{(3)} - \mathbf{o}}{\|\bar{\mathbf{r}}_m^{(3)} - \mathbf{o}\|_2}. \tag{16}$$

Candidate reflector positions are sampled along this ray. For each candidate position, the reflection likelihood is evaluated using the reflective surface probability map $\hat{\mathbf{S}}_t$. The reflector geometry $(\mathbf{c}_m, \mathbf{n}_m)$ is selected as the candidate that maximizes the accumulated reflection consistency.

*Geometric Mirroring.* Given the estimated $(\mathbf{c}_m, \mathbf{n}_m)$ and $\bar{\mathbf{r}}_m^{(3)}$, the NLOS pedestrian position is recovered via geometric mirroring with respect to the reflective surface. The recovered position $\hat{\mathbf{x}}^{(3)}$ is computed as

$$\hat{\mathbf{x}}_m^{(3)} = \bar{\mathbf{r}}_m^{(3)} - 2\big((\bar{\mathbf{r}}_m^{(3)} - \mathbf{c}_m) \cdot \mathbf{n}_m\big)\mathbf{n}_m. \tag{17}$$

This operation corresponds to the geometric reflection of $\bar{\mathbf{r}}_m^{(3)}$ across the estimated planar reflector, consistent with third-order reflection propagation.

*Final Localization.* The final set of object location candidates at time $t$ is constructed by combining the first-order reflection cluster centers with the mirrored third-order estimates:

$$\hat{X}_t = \{\bar{\mathbf{r}}_k^{(1)}\}_k \, \cup \, \{\hat{\mathbf{x}}_m^{(3)}\}_m. \tag{18}$$

Spatially adjacent candidates are merged to obtain a refined object location estimate. This physics-guided inference procedure leverages the predicted reflection types and the estimated reflective surface geometry to enable reliable NLOS pedestrian localization without requiring prior scene structure information.

## 5 Dataset

The dataset was collected in a constructed urban intersection testbed designed to capture NLOS pedestrian scenarios. The ego vehicle is equipped with a front-view RGB camera, a 77 GHz mmWave radar, a LiDAR, and a wheel encoder. To obtain accurate ground-truth for occluded pedestrians, an overhead camera was installed above the center of the intersection to capture a top-down view of the entire scene. All sensor data were synchronized using ROS. Radar data were acquired using a TI AWR2944EVM sensor [30]. Raw radar signals were processed into radar PCD following the methodology described in [19]. The radar PCD generation itself is not part of the proposed contribution.

The dataset consists of scenarios where one to three pedestrians move in an intersection environment. Pedestrians appear from occluded side roads and move across the intersection while the ego vehicle either remains stationary or moves through the intersection. Two spatial configurations are defined based on the intersection geometry. B1 corresponds to the case where a frontal wall blocks the opposite side of the intersection, creating strong NLOS conditions. B2 represents the configuration where the opposite road is open, allowing the ego vehicle to drive straight or turn right. Both ego-static and ego-dynamic driving conditions are included in the dataset. In the ego-dynamic setting, the ego vehicle drives through the testbed at speeds up to 22.00 km/h, and the evaluation covers NLOS pedestrians with a radar-based NLOS observation range of up to 17.92 m. In total, the dataset contains 44 training scenarios and 13 validation scenarios under ego-static conditions, and 57 training scenarios with 10 evaluation scenarios under ego-dynamic conditions. Detailed dataset information is provided in the Supplementary material.

## 6 Experimental Results

Experiments are conducted using the dataset described in Section 5. Evaluation is performed under both ego-static and ego-dynamic conditions across the two spatial configurations B1 and B2. The proposed method is evaluated from three perspectives: reflection-type point segmentation, reflective surface estimation, and final NLOS pedestrian localization. Qualitative results are illustrated in Fig. 4. Detailed experimental results are provided in the Supplementary material.

For a fair comparison, all radar-based baselines and our method are evaluated using the same ego-motion-compensated radar PCD, which is generated by FMCW processing, CFAR detection, clustering, coordinate transformation, and wheel-encoder-based motion compensation. Thus, motion compensation is applied consistently to all compared methods rather than only to the proposed method. All methods are evaluated under the same ego-static or ego-dynamic protocol, association rule, and average localization-error metric.

### 6.1 Reflection-Type Point Segmentation Performance

We evaluate the proposed reflection-type point segmentation on the dataset defined in Section 5 to assess its ability to identify informative radar points for NLOS localization. The task focuses on distinguishing pedestrian-originated reflections from environmental reflections and noise.

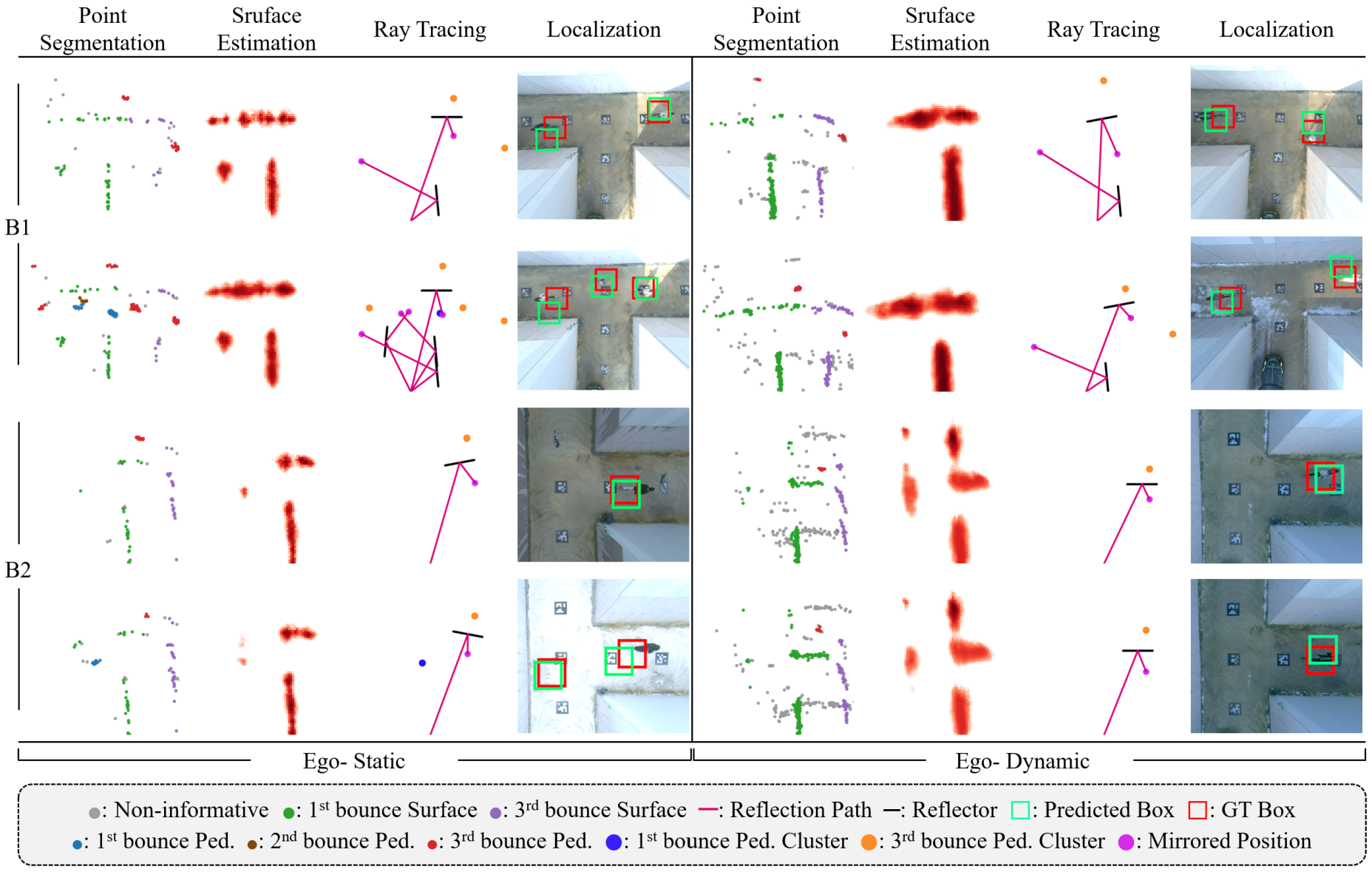

**Fig. 4: Qualitative results of the proposed reflection-aware NLOS localization framework under ego-static and ego-dynamic settings**. Columns show intermediate stages of the pipeline: point segmentation, reflective surface estimation, ray tracing, and final localization. The final column visualizes ground-truth and predicted pedestrian locations.

Performance is evaluated using class-wise precision, recall, and F1-score, and compared with baseline methods [14,34]. Experiments are conducted under both ego-static and ego-dynamic conditions.

The proposed method achieves 94.6% accuracy and a macro-F1 of 0.915 in ego-static settings, and 89.9% accuracy with a macro-F1 of 0.814 in ego-dynamic settings, as summarized in Table 2. Since NLOS pedestrians are observed through third-order reflections, accurate classification of third-bounce points is particularly important. The proposed approach maintains strong performance for third-bounce classification in both ego-static and ego-dynamic conditions.

Compared to baseline methods [14, 34], the proposed method consistently improves overall segmentation performance as well as third-bounce classification accuracy, demonstrating its effectiveness in isolating radar observations relevant to NLOS pedestrians.

**Table 2: Point-wise reflection path classification performance**. Overall results are reported using Accuracy and Macro-F1, while precision, recall, and F1-score are additionally reported for the NLOS pedestrian reflection class corresponding to third-bounce reflections, which are critical for NLOS position reconstruction.

| Method | Scenario | Accuracy | Macro-F1 | 3rd-bounce pedestrian | | |
|---|---|---|---|---|---|---|
| | | | | Precision | Recall | F1-score |
| Kraus et al. [14] | Ego-static | 0.686 | 0.661 | 0.757 | 0.639 | 0.693 |
| | Ego-dynamic | 0.677 | 0.635 | 0.553 | 0.863 | 0.674 |
| Wu et al. [34] | Ego-static | 0.930 | 0.890 | 0.884 | 0.931 | 0.907 |
| | Ego-dynamic | 0.610 | 0.481 | 0.250 | 0.501 | 0.334 |
| Proposed | Ego-static | 0.946 | 0.915 | 0.897 | 0.935 | 0.916 |
| | Ego-dynamic | 0.899 | 0.814 | 0.744 | 0.718 | 0.731 |

**Table 3: Quantitative evaluation of reflective surface estimation**.

| Ego Status | Scenario | RPE (m) | RAE (deg) |
|---|---|---|---|
| Static | B1 | 0.280 | 7.175 |
| | B2 | 0.415 | 6.698 |
| Dynamic | B1 | 0.285 | 7.207 |
| | B2 | 0.357 | 5.997 |

**Table 4: System performance**.

| Metric | Value |
|---|---|
| Reflection-aware learning | 55.22 ms |
| Physics-guided ray tracing | 16.80 ms |
| FPS | 13.85 |
| Early detection | 0.77 s |

### 6.2 Reflective Surface Estimation Performance

We evaluate the accuracy of the inferred reflective surfaces, which are used for NLoS object localization. Quantitative evaluation is performed using pseudo ground-truth wall segments extracted from LiDAR.

Given that the dataset environment consists of linear walls in a T-junction, reflective surfaces are approximated as line segments using DBSCAN clustering and RANSAC-based line fitting on LiDAR point clouds. The inferred reflector geometry is then compared against these LiDAR-derived pseudo labels.

To assess geometric consistency, we measure both reflection point error (RPE) and reflection angle error (RAE).

RPE is defined as the Euclidean distance between the predicted reflection point $\mathbf{p}_{\text{pred}}$ and the LiDAR-derived ground-truth reflection point $\mathbf{p}_{\text{gt}}$:

$$\text{RPE} = \|\mathbf{p}_{\text{pred}} - \mathbf{p}_{\text{gt}}\|_2 . \tag{19}$$

Here, $\mathbf{p}_{\text{gt}}$ is defined as the intersection between the ray connecting the radar origin and the third-order cluster center and the LiDAR-extracted reflector. $\mathbf{p}_{\text{pred}}$ is defined analogously using the predicted reflector.

**Table 5: Quantitative comparison of pedestrian localization performance.** Results are reported in terms of average localization error (AE, m).

| Ego Vehicle Status | Static | | Dynamic | |
|---|---|---|---|---|
| Environments | LOS (m) | NLOS (m) | LOS (m) | NLOS (m) |
| Scheiner et al. [27] | 1.02 | 2.36 | 1.36 | 4.74 |
| Park et al. [24] | 0.89 | 1.69 | 1.41 | 3.23 |
| Proposed | 0.44 | 1.01 | 0.54 | 1.23 |

RAE measures the angular difference between the predicted reflector direction $\mathbf{d}_{\text{pred}}$ and the LiDAR-derived reflector direction $\mathbf{d}_{\text{gt}}$:

$$\text{RAE} = \arccos\left(\left|\frac{\mathbf{d}_{\text{pred}} \cdot \mathbf{d}_{\text{gt}}}{\|\mathbf{d}_{\text{pred}}\| \, \|\mathbf{d}_{\text{gt}}\|}\right|\right) \times \frac{180}{\pi}. \tag{20}$$

Quantitative results are summarized in Table 3. Under ego-static conditions, the proposed method achieves an RPE of 0.280 m in intersection layout B1 and 0.415 m in layout B2, with corresponding RAEs of 7.175° and 6.698°. Under ego-dynamic conditions, the RPE remains 0.285 m in layout B1 and 0.357 m in layout B2, while the RAE is 7.207° and 5.997°, respectively. These results demonstrate consistent sub-meter reflection point accuracy and reflection angle errors below 8° across different spatial configurations under both ego-static and ego-dynamic conditions.

### 6.3 Object Localization Performance

Localization accuracy is evaluated using the average localization error, defined as the mean localization distance described in Eq. 3. Evaluation is conducted under both ego-static and ego-dynamic conditions, with results reported separately for LOS and NLOS pedestrian localization.

As summarized in Table 5, the proposed method achieves an average localization error of 0.44 m for LOS pedestrians and 1.01 m for non-line-of-sight pedestrians under ego-static conditions. Under ego-dynamic conditions, the error is 0.54 m for LOS pedestrians and 1.23 m for NLOS pedestrians.

Compared with geometry-based baselines, the proposed approach substantially improves NLOS pedestrian localization accuracy. Under ego-static conditions, the NLOS localization error decreases from 1.69 m to 1.01 m, corresponding to an improvement of approximately 40%. Under ego-dynamic conditions, the error decreases from 3.23 m to 1.23 m, yielding a reduction of more than 60%. The increase in NLOS localization error from ego-static to ego-dynamic conditions remains limited, indicating robustness against motion-induced disturbances.

Finally, Table 4 reports the runtime performance of the proposed framework. The reflection-aware learning stage takes 55.22 ms per frame, and the physics-guided ray-tracing module takes 16.80 ms, resulting in an overall throughput of 13.85 FPS on an RTX 3090 GPU. The model is lightweight, with 1.17M parameters and 55.04 GMACs per frame, corresponding to 110.08 GFLOPs when one multiply–accumulate operation is counted as two floating-point operations. The semantic auxiliary head is used only during training and therefore introduces no additional inference cost.

## 7 Conclusion & Discussion

This paper presented a reflection-aware learning framework for NLOS pedestrian localization in intersection scenarios. The proposed method jointly learns reflection-type point segmentation and reflective surface estimation, and integrates them with a physics-guided ray tracing module for structured multipath interpretation. This formulation enables data-driven decomposition of multipath signals while preserving physical consistency.

Experimental results demonstrate improved NLOS localization accuracy under both ego-static and ego-dynamic conditions. The proposed framework detects NLOS pedestrians on average 0.77 s earlier than first-bounce radar observations by leveraging third-order multipath reflections. The system operates near real time, achieving 72 ms end-to-end latency corresponding to 13.85 FPS.

Further analysis shows that accurate reflection-point detection alone is insufficient for reliable NLOS pedestrian inference. Localization error increases significantly when reflective surface estimation becomes inaccurate, indicating a strong interdependence between reflection-type segmentation and reflective surface reasoning.

Overall, the results highlight that combining reflection-aware representation learning with physics-guided reasoning enables reliable and early NLOS perception beyond direct LOS.

## Acknowledgements

This work was supported by Seoul National University, including the provision of experimental equipment and the construction of the testbed environment for data collection. This work was also supported by the Korea Institute for Advancement of Technology (KIAT) grant funded by the Ministry of Trade, Industry and Energy (MOTIE) (P0020536, HRD Program for Industrial Innovation), a National Research Foundation of Korea (NRF) grant funded by the Ministry of Science and ICT (MSIT) (No. RS-2026-25518893), and a research grant from LG Electronics. This research was also supported by the Advanced GPU Utilization Support Program funded by the Ministry of Science and ICT (MSIT), Republic of Korea.

## 1 Dataset Details

The goal of this study is to analyze the propagation paths observed in 2D radar point clouds (PCD) in order to estimate the positions of non-line-of-sight (NLOS) pedestrians and ultimately contribute to collision avoidance. To this end, the dataset was collected in a real-world environment using a dedicated testbed designed to emulate narrow urban roads. The testbed spans approximately 53.5 m $\times$ 33.5 m and includes road segments with a width of approximately 5.5 m, as well as T-junction structures and occlusion regions that generate realistic NLOS conditions.

Using this testbed environment, a large-scale dataset was collected to capture diverse pedestrian–vehicle interaction scenarios. The dataset consists of 120 scenarios in total: 67 ego-dynamic scenarios and 53 ego-static scenarios. Specifically, ego-dynamic refers to situations where the ego-vehicle is actively driving, while ego-static denotes cases where the ego-vehicle is completely stopped. In total, the dataset provides 12,539 frames and 2,853,228 radar points.

Within these scenarios, several factors were systematically varied in order to represent realistic urban driving conditions. Each scenario differs in the number of pedestrians, pedestrian motion direction, vehicle speed, and the spatial configuration of the environment. The number of pedestrians per scenario ranges from one to three. Among the 120 scenarios, 32 scenarios contain one pedestrian, 59 scenarios contain two pedestrians, and 29 scenarios contain three pedestrians.

Pedestrian motion directions were designed to reflect realistic alleyway and intersection situations in which line-of-sight (LOS) and NLOS pedestrians may coexist. LOS pedestrians either approach the ego vehicle or move away from it, while NLOS pedestrians approach the ego vehicle from the left or right branches of the intersection. The detailed scenario configurations are summarized in Table 2. Each configuration specifies the number of LOS pedestrians approaching or leaving the ego vehicle, the number of NLOS pedestrians appearing from the left or right branches, and the total number of pedestrians in the scene.

Two intersection layouts are considered in the dataset. B1 corresponds to the case where a frontal wall blocks the opposite side of the intersection, creating strong NLOS conditions. B2 represents the configuration where the opposite road is open, allowing the ego vehicle to drive straight or turn right. Both ego-static and ego-dynamic driving conditions are included in the dataset.

To capture multimodal observations of these scenarios, the ego-vehicle was equipped with multiple sensors. The dataset includes radar, camera, LiDAR, wheel encoder, and a top-down view camera used for ground-truth acquisition. As the radar sensor, we used the TI AWR2944EVM, a 77 GHz automotive mmWave FMCW radar, and ADC data were collected at 10 FPS through the TI DCA1000 module. The radar provides a maximum detection range of 17.92 m and a range resolution of 7 cm. In addition to the radar, the ego-vehicle is equipped with a 12-megapixel camera with a fisheye lens, a 128-channel LiDAR sensor, and wheel encoders for vehicle motion measurement. All sensors operate at 10 Hz and are mounted on the ego-vehicle to capture synchronized multimodal observations of the surrounding environment.

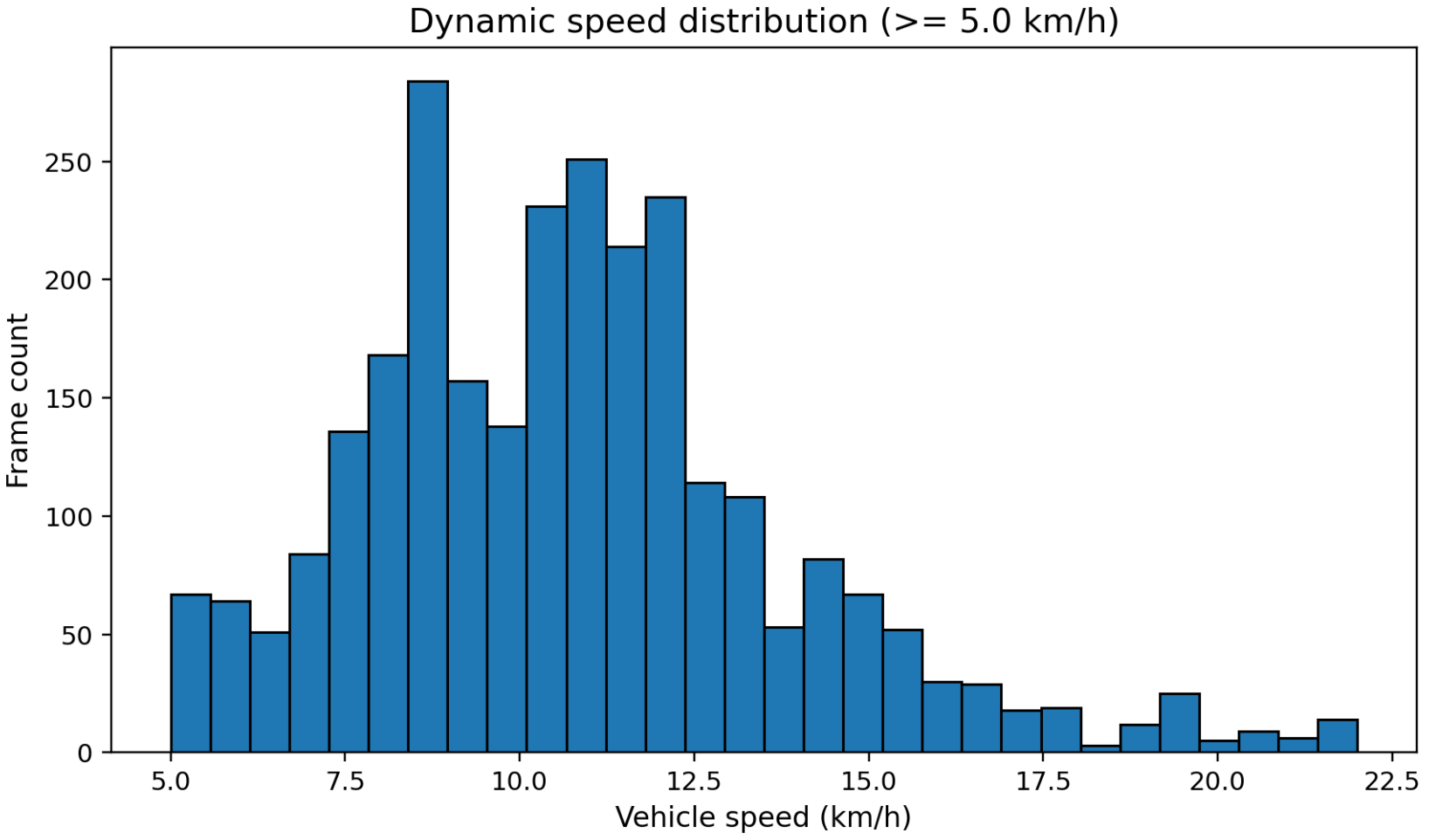

**Fig. 1: Vehicle speed distribution in ego-dynamic scenarios.** The histogram shows the distribution of vehicle speeds measured using wheel encoders for frames with speeds greater than or equal to 5 km/h.

Finally, the vehicle motion in ego-dynamic scenarios was designed to reflect realistic low-speed driving conditions in narrow urban roads. The vehicle motion was measured using wheel encoders mounted on the ego vehicle. To exclude near-static frames, only dynamic frames with vehicle speeds greater than or equal to 5 km/h were considered for speed analysis. Under this protocol, the average speed is 10.78 km/h with a standard deviation of 3.06 km/h Fig. 1. The maximum speeds is 22.00 km/h, while the scenario-level mean speed ranges from 7.33 km/h to 21.30 km/h.

**Table 1: Overall statistics of the collected dataset.** The scenarios are categorized by the number of pedestrians and the ego-vehicle's driving conditions, demonstrating the structural diversity of the NLOS interactions.

| Pedestrian Count | Ego-Static | Ego-Dynamic | Total Scenarios |
|---|---|---|---|
| 1 Pedestrian | 14 | 18 | 32 |
| 2 Pedestrians | 25 | 34 | 59 |
| 3 Pedestrians | 14 | 15 | 29 |
| **Total** | **53** | **67** | **120** |

**Table 2: Scenario configurations used in the dataset.** Each row represents a predefined combination of LOS and NLOS pedestrian behaviors. The columns indicate the number of pedestrians approaching or leaving the ego-vehicle in LOS, the number of pedestrians appearing from the left or right NLOS branches, and the number of scenarios instantiated for each configuration.

| **Intersection** | **LOS (approach)** | **LOS (leave)** | **NLOS (left)** | **NLOS (right)** | **Scenarios** |
|---|---|---|---|---|---|
| B1 | | | | 1 | 10 |
| B1 | | | 1 | | 11 |
| B1 | | | | 2 | 11 |
| B1 | | | 1 | 1 | 8 |
| B1 | | | 2 | | 12 |
| B1 | | 1 | | 1 | 4 |
| B1 | | 1 | 1 | | 3 |
| B1 | 1 | | | 1 | 3 |
| B1 | 1 | | 1 | | 4 |
| B1 | | | 1 | 2 | 6 |
| B1 | | | 2 | 1 | 8 |
| B1 | | 1 | | 2 | 4 |
| B1 | | 1 | 2 | | 4 |
| B1 | 1 | | | 2 | 4 |
| B1 | 1 | | 2 | | 3 |
| B2 | | | | 1 | 11 |
| B2 | | | 2 | | 10 |
| B2 | | 1 | | 1 | 2 |
| B2 | 1 | | | 1 | 2 |
| **Total** | **5** | **5** | **15** | **16** | **120** |

### 1.1 Experiment Protocol

This section provides additional details on the evaluation protocol used for the radar-based baselines and our method. All radar-based methods, including the baselines [1–4] and the proposed method, are evaluated using the same processed radar PCD. The radar PCD is generated through FMCW signal processing, CFAR detection, clustering, coordinate transformation, and ego-motion compensation based on wheel-encoder measurements. Thus, the input radar PCD is consistently preprocessed across all compared methods.

For methods with publicly available implementations, we use the official code when available and adapt only the input interface to accept the processed radar PCD. For [3], which relies on reflector geometry, LiDAR-derived reflector segments are provided to construct the required geometric input. For [2], we follow its rule-based geometric reasoning procedure using the same processed PCD and evaluation setting. All methods are evaluated under the same ego-static and ego-dynamic split, matching rule, and localization-error metric.

**Table 3: Radar PCD quality under ego-static and ego-dynamic conditions.**

| Metric | Ego-static | Ego-dynamic |
|---|---|---|
| Clutter/unlabeled ratio | 13.8% | 31.6% |
| Surface coverage | 0.478 | 0.412 |
| Surface concentration | 0.135 | 0.213 |

**Table 4: Ablation studies under the ego-dynamic setting.**

| Variant | Point Acc. | Macro-F1 | Surf. IoU |
|---|---|---|---|
| Proposed | 0.898 | 0.818 | 0.738 |
| w/o Camera | 0.884 | 0.800 | 0.672 |
| w/o Aux loss | 0.896 | 0.819 | 0.726 |

Table 3 summarizes the radar PCD quality under ego-static and ego-dynamic conditions. Compared with the ego-static setting, the ego-dynamic setting shows a higher clutter/unlabeled ratio, lower surface coverage, and higher surface concentration. Specifically, the clutter/unlabeled ratio increases from 13.8% to 31.6%, surface coverage decreases from 0.478 to 0.412, and surface concentration increases from 0.135 to 0.213.

These results indicate that ego-dynamic radar PCDs are more challenging even after ego-motion compensation. Ego motion affects not only coordinate alignment but also the spatial distribution and reliability of radar returns. In particular, useful surface reflections become sparser, reflector observations are weakened, and the remaining returns become more spatially concentrated. These properties make ego-dynamic scenes more difficult for methods that rely heavily on explicit geometric reasoning or direct point-wise input distributions.

Table 4 reports the ablation results under the ego-dynamic setting. Removing the camera input decreases the point accuracy from 0.898 to 0.884, the Macro-F1 score from 0.818 to 0.800, and the surface IoU from 0.738 to 0.672. This result shows that camera information provides complementary scene context, especially for estimating surface-level structures from sparse and noisy radar observations.

Removing the auxiliary loss slightly changes the point-level metrics but reduces the surface IoU from 0.738 to 0.726. This suggests that the auxiliary supervision contributes to learning a more consistent reflection-aware representation, which is particularly helpful for preserving surface-level structure under degraded ego-dynamic radar PCD conditions.

### 1.2 Image Segmentation Annotation

Front-view camera images were manually annotated with pixel-level semantic segmentation masks to provide precise spatial context. We focused on two key classes for analyzing NLOS interactions: *wall* for structural occlusion boundaries, and *pedestrian* for dynamic targets. Labels were created at a 2-frame interval, corresponding to a frequency of 5 Hz. As shown in Fig. 2, all annotations were performed using CVAT to maintain high accuracy, successfully capturing fine details even in challenging scenarios with severe partial occlusions.

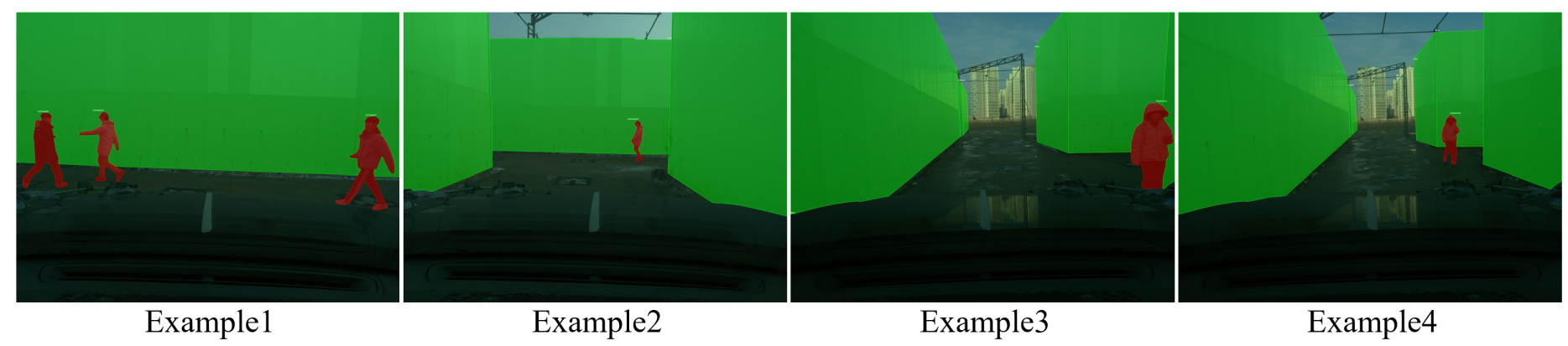

**Fig. 2: Examples of semantic segmentation annotations.** The dataset provides pixel-level masks for structural boundaries labeled as *wall* and dynamic targets labeled as *pedestrian*. The manual annotations maintain high accuracy even in highly occluded, NLOS conditions, providing robust ground-truth data for complex urban interaction scenarios.

**Table 5: Distribution of radar point classes in the dataset.**

| Point class | Points | Ratio |
|---|---|---|
| None | 708,343 | 24.83% |
| 1st bounce | 584,611 | 20.49% |
| 2nd bounce | 89,056 | 3.12% |
| 3rd bounce | 117,835 | 4.13% |
| 1st bounce surface | 1,032,335 | 36.18% |
| 3rd bounce surface | 321,048 | 11.25% |
| Total | 2,853,228 | 100% |

### 1.3 Point Class Annotation

Building upon the reflection path definitions established in the main text, each radar point was assigned a semantic class based on its propagation order and physical origin. We defined six specific point classes: *None*, *1st bounce*, *2nd bounce*, *3rd bounce*, *1st bounce surface*, and *3rd bounce surface*.

Specifically, the classes *1st bounce*, *2nd bounce*, and *3rd bounce* correspond to object-originated reflections generated by the pedestrians. Conversely, *1st bounce surface* and *3rd bounce surface* denote surface-originated reflections generated by the structural walls, which actively contribute to the multipath propagation. All remaining points that do not belong to these specific paths were labeled as *None*.

To perform the labeling, the geometric structure of each intersection environment was first reconstructed using LiDAR point clouds. Specifically, DBSCAN and RANSAC algorithms were applied to the LiDAR points to extract wall structures, which were represented as a set of linear reflector segments. These reflectors were then used to analyze radar propagation paths. Next, virtual rays were generated from the radar origin to the annotated pedestrian positions in order to determine pedestrian visibility and possible reflection paths. The distinction between LOS and NLOS pedestrians was determined using the ray connecting the radar origin $(0, 0)$ and the pedestrian location. If this ray did not intersect any reflector, the pedestrian was considered LOS; otherwise, the pedestrian was classified as NLOS.

The *1st bounce surface* class was assigned to static radar points located near reflective surfaces. For each point, the shortest distance to the extracted reflectors was computed. If the distance to the nearest reflector was less than or equal to 0.50 m, the point was considered a candidate surface point. Then, the first reflector intersected by the ray from the radar origin to the point was identified. If this reflector coincided with the nearest reflector, the point was labeled as *1st bounce surface*. Otherwise, the point was regarded as occluded by another reflector and the label was not assigned.

The *3rd bounce surface* class was assigned to static points that were not labeled as *1st bounce surface*. Only points whose ray from the radar origin intersected at least one reflector were considered. For each such point, the first intersected reflector was used to generate a mirrored point through reflection. The distance between the mirrored point and other reflectors was then evaluated. If the mirrored point was located within 0.50 m of another reflector, the corresponding radar point was labeled as *3rd bounce surface*. Cases where the mirrored point corresponded to the same reflector used for the reflection were excluded in order to ensure that only valid multipath reflections were considered.

Finally, the classes *1st bounce*, *2nd bounce*, and *3rd bounce* were assigned based on the propagation path order derived from virtual ray tracing between the radar origin, pedestrians, and reflectors. Points associated with reflective surfaces were labeled as surface classes, while points corresponding to pedestrian reflection paths were labeled as bounce classes. Points that did not satisfy any of these conditions were labeled as *None*.

Across the entire dataset, 708,343 points correspond to the *None* class, 584,611 points correspond to *1st bounce*, 89,056 points correspond to *2nd bounce*, 117,835 points correspond to *3rd bounce*, 1,032,335 points correspond to *1st bounce surface*, and 321,048 points correspond to *3rd bounce surface* Table 5.

## 2 Additional Experimental Results

### 2.1 Qualitative Results

Fig. 3a and Fig. 3b present representative qualitative examples of the proposed method under ego-static and ego-dynamic scenarios, respectively. Each figure includes four example scenes illustrating the estimated NLOS pedestrian locations.

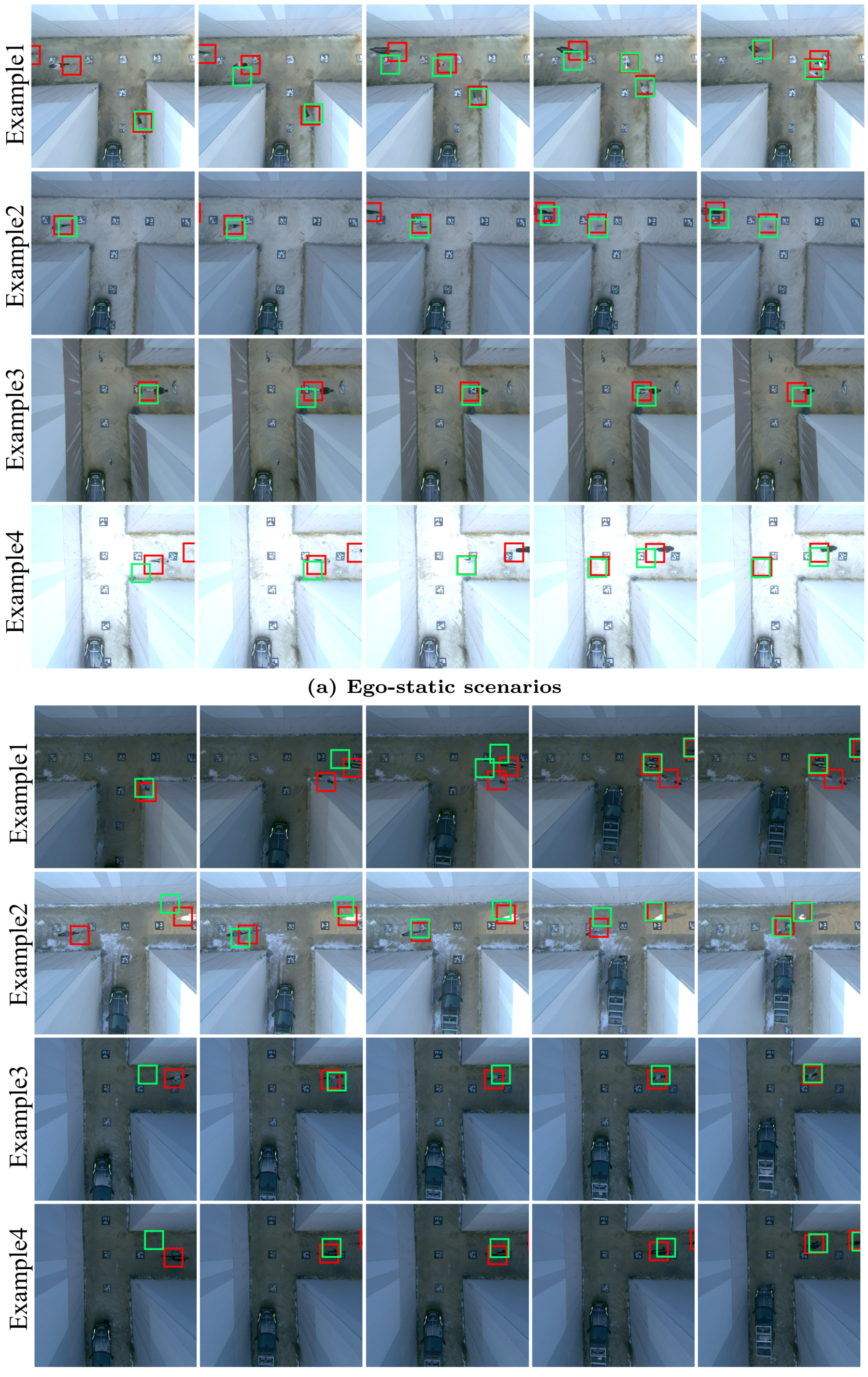

**(a) Ego-static scenarios**

**(b) Ego-dynamic scenarios**

**Fig. 3: Qualitative results in ego-centric scenarios.** (a) and (b) show the results in ego-static and ego-dynamic scenarios, respectively. The green bounding boxes indicate the predicted NLOS pedestrian locations, while the red bounding boxes denote the ground truth.

**Table 6:** Ablation study on query–key–value modality assignments. Different radar and camera feature assignments are evaluated in the cross-attention module. The table reports point-level classification performance and reflective-surface heatmap prediction quality. For surface prediction, lower MAE and higher Dice indicate better performance.

| Cross-attention modality | | | Point classification | | | Surface heatmap | |
|---|---|---|---|---|---|---|---|
| Query | Key | Value | Acc. | Prec. | Rec. | MAE↓ | Dice↑ |
| Camera | Camera | Radar | 0.8443 | 0.7403 | 0.7383 | 0.1027 | 0.5046 |
| Camera | Radar | Camera | 0.8433 | 0.7339 | 0.7198 | 0.1027 | 0.5187 |
| Camera | Radar | Radar | 0.8218 | 0.7273 | 0.7223 | 0.1053 | 0.4692 |
| Radar | Camera | Camera | 0.8550 | 0.7612 | 0.7102 | 0.1000 | 0.5473 |
| Radar | Camera | Radar | 0.8563 | 0.7613 | 0.7429 | 0.1008 | 0.5191 |
| Radar | Radar | Camera | 0.8616 | 0.7610 | 0.7497 | 0.1005 | 0.5232 |
| Radar | Radar | Radar | 0.8451 | 0.7621 | 0.7211 | 0.1043 | 0.4800 |

### 2.2 Ablation Study on Q–K–V Modality Assignments

To analyze the contribution of different modality assignments in the proposed cross-attention module, we conduct an ablation study by varying the input modalities used for the query (Q), key (K), and value (V). Specifically, radar and camera features are assigned to Q, K, and V in different combinations, and the resulting model performance is evaluated.

For each configuration, we measure point-level classification performance using accuracy, precision, and recall. In addition, we evaluate the quality of reflective surface heatmap prediction using mean absolute error (MAE) and Dice score. The MAE metric measures the average difference between the predicted heatmap and the ground truth heatmap, while the Dice score evaluates the spatial overlap between the predicted and ground truth surface regions after thresholding.

Table 6 summarizes the results of the ablation study. The results show that using radar features as the query generally leads to improved point classification performance. In particular, the radar–radar–camera configuration achieves the highest point accuracy and recall, suggesting that radar features are well suited for query representations in the cross-attention mechanism.

On the other hand, the radar–camera–camera configuration achieves the lowest surface MAE and the highest Dice score for reflective surface prediction. This observation indicates that camera features used as key and value provide useful structural cues for estimating reflective surfaces.

Note that the ablation experiments focus on analyzing the relative performance differences among modality configurations. To reduce computational cost, these experiments are conducted using a simplified training setup rather than the full training configuration used in the main experiments. Therefore, the reported results primarily highlight relative trends among different modality assignments rather than absolute performance values.

**Table 7: Localization error of the fixed-geometry variant.** LiDAR-derived ground-truth wall geometry is used instead of the learned reflective-surface estimation.

| Setting | LOS Error (m) | NLOS Error (m) |
|---|---|---|
| Ego-static | 0.375 | 0.840 |
| Ego-dynamic | 0.535 | 1.230 |

### 2.3 Camera and Geometry Ablations

We conduct additional ablation studies to analyze the effects of camera input and reflective-surface geometry. First, we evaluate a camera-off variant under the ego-dynamic setting by removing the visual BEV context from the proposed model. As shown in Table 4, removing the camera input decreases the point accuracy from 0.898 to 0.884, the Macro-F1 score from 0.818 to 0.800, and the surface IoU from 0.738 to 0.672. These results indicate that camera-derived BEV context provides complementary spatial cues for radar-based NLOS reasoning. In particular, visual context helps improve both reflection-order reasoning and reflective-surface estimation from sparse and noisy radar observations.

We further evaluate a fixed-geometry variant, where the learned reflective-surface estimation is replaced with LiDAR-derived ground-truth wall geometry. This setting provides a reference for analyzing the effect of accurate reflector geometry on localization performance. As shown in Table 7, the fixed-geometry variant achieves LOS localization errors of 0.375 m and 0.535 m, and NLOS localization errors of 0.840 m and 1.230 m under ego-static and ego-dynamic settings, respectively. These results show that accurate geometry is beneficial for NLOS localization. However, unlike this fixed-geometry variant, the proposed method estimates reflective surfaces directly from camera-radar input and does not require additional LiDAR-derived reflector geometry at test time.

### 2.4 Semantic Auxiliary Loss and Model Complexity

We also analyze the contribution of the semantic auxiliary loss. As shown in Table 4, removing the auxiliary loss keeps the point-level metrics nearly unchanged, with point accuracy changing from 0.898 to 0.896 and Macro-F1 from 0.818 to 0.819. However, the surface IoU decreases from 0.738 to 0.726. This suggests that the semantic auxiliary loss mainly improves the consistency and generalization of reflective-surface estimation rather than directly improving point-wise classification performance.

The auxiliary supervision encourages the model to learn a more structured reflection-aware representation, which is helpful for preserving surface-level geometry under degraded ego-dynamic radar PCD conditions. Therefore, although the semantic auxiliary loss has a limited effect on point-level classification accuracy, it contributes to more stable reflective-surface estimation and improves the robustness of the learned representation.

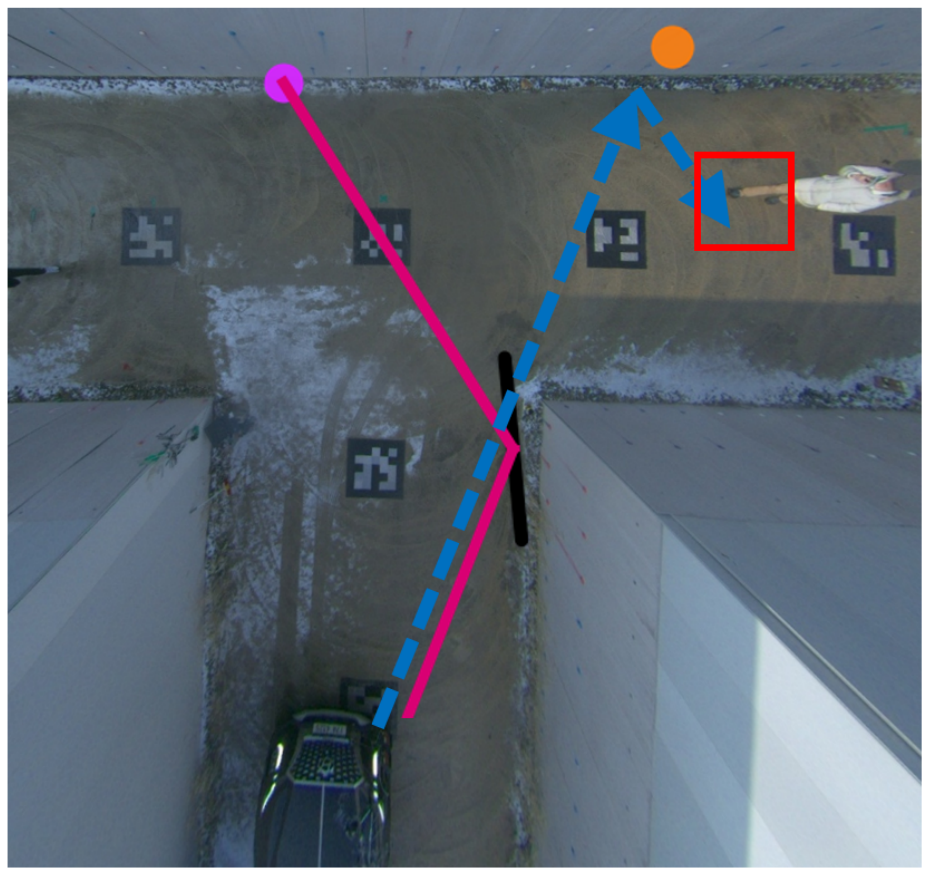
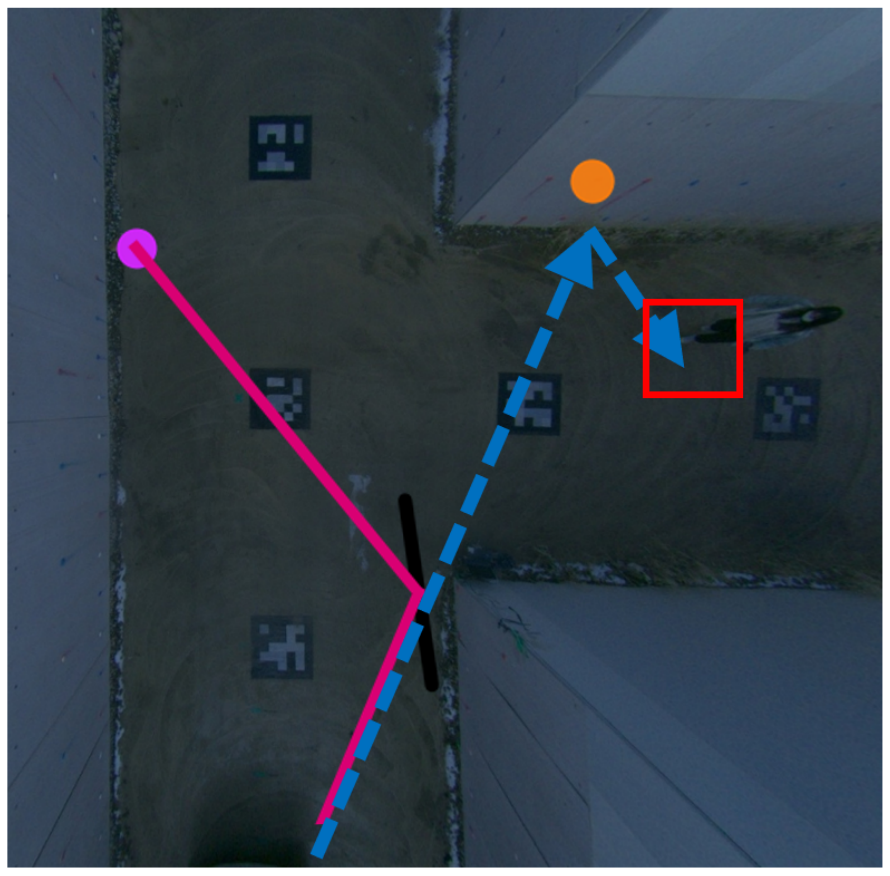

**Fig. 4: Representative failure cases.** The red bounding box indicates the ground-truth location of the NLOS pedestrian. The blue dashed arrows illustrate the correct multipath propagation derived from the actual scene geometry. However, due to the geometric ambiguity of the corner reflector indicated by the black segment, the model incorrectly infers the reflection path, represented by the solid pink line. Consequently, this failure in spatial reasoning leads the model to predict the pedestrian at an entirely incorrect location at the end of the pink path rather than the true position.

### 2.5 Failure Case

While our model successfully estimates the positions of NLOS pedestrians in most scenarios, we observed certain failure cases when the spatial reasoning of multipath propagation is disrupted. As shown in Fig. 4, severe estimation errors occur when the model misinterprets the complex geometric structure of the intersection. In such cases, the highly ambiguous radar points cause the model to output bounding boxes with significant positional offsets. This highlights the inherent challenge of spatial reasoning in extreme NLOS conditions, which we leave as an area for future improvement.